\documentclass[12pt]{article}
\usepackage{lscape}
\usepackage{geometry}
\usepackage[onehalfspacing]{setspace}
\usepackage{amssymb}
\usepackage{amsfonts}
\usepackage{graphicx}
\usepackage{amsmath}
\usepackage{epstopdf}
\usepackage{natbib}
\usepackage{bibunits}
\usepackage{morefloats}
\usepackage{float}
\usepackage{comment}
\usepackage{pdflscape}
\usepackage{adjustbox}
\usepackage{subcaption}
\usepackage{longtable}
\usepackage{apalike}
\usepackage{xfrac}
\usepackage[dvipsnames,table]{xcolor}
\usepackage{etoolbox,siunitx}
\robustify\bfseries

\usepackage{hhline,multirow}

\usepackage{qtree}
\usepackage{multicol}
\usepackage{booktabs}
\usepackage{threeparttable}
\usepackage{tabularx}
\usepackage{dcolumn}
\newcolumntype{d}[1]{D..{#1}} 

\usepackage{siunitx}
\usepackage{mathpazo} 
\usepackage[scaled]{helvet} 
\usepackage{courier} 
\normalfont
\usepackage[T1]{fontenc}
\usepackage{listings}
\usepackage{epigraph}
\def\sym#1{\ifmmode^{#1}\else\(^{#1}\)\fi}

\usepackage[pdftex,bookmarks,colorlinks]{hyperref}
\hypersetup{
  colorlinks,
  citecolor=PineGreen,
  linkcolor=red,
  urlcolor=blue}
\usepackage{cleveref}

\DeclareMathOperator*{\argmin}{arg\,min}

\usepackage{etoolbox}

\makeatletter
\patchcmd{\epigraph}{\@epitext{#1}}{\itshape\@epitext{#1}}{}{}
\makeatother

\renewenvironment{abstract}
 {\small
  \begin{center}
  \bfseries \abstractname\vspace{-.5em}\vspace{0pt}
  \end{center}
  \list{}{
    \setlength{\leftmargin}{2.05cm}    \setlength{\rightmargin}{\leftmargin}  }  \item\relax}
 {\endlist}
\usepackage{graphicx}
\usepackage{wrapfig}

\usepackage{algorithm,algorithmic}

\begin{document}
\sloppy

\title{\vspace*{-0cm} \Huge Slow-Growing Trees} 
\author{Philippe Goulet Coulombe\thanks{%
Departement des Sciences Économiques, \href{mailto:p.gouletcoulombe@gmail.com}{{p.gouletcoulombe@gmail.com}}. For helpful discussions and/or  comments, I thank Karun Adusumilli, Maximilian G{\"o}bel, David Wigglesworth and Boyuan Zhang.}}
\date{\vspace{-0.4cm}
Université du Québec à Montréal \\[2ex]%
\small
\small
First Draft: February 4  2021 \\
This Draft: \today \\ 
\href{https://drive.google.com/file/d/1W_-t5EOn3ulLz6SVOVs1AHHGrc7SAb4h/view?usp=sharing}{Latest Draft Here} \\ 
\vspace{0.3cm}
\large
  }
\maketitle


\begin{abstract}
\noindent Random Forest's performance can be matched by a single slow-growing tree (SGT), which uses a  learning rate to tame CART's greedy algorithm. SGT exploits the view that CART is an extreme case of an iterative weighted least square procedure. Moreover, a unifying view of Boosted Trees (BT) and Random Forests (RF) is presented. Greedy ML algorithms' outcomes can be improved using either "slow learning" or diversification. SGT applies the former to estimate a single deep tree, and \textit{Booging} (bagging stochastic BT with a high learning rate) uses the latter with additive shallow trees. The performance of this tree ensemble quaternity (Booging, BT, SGT, RF) is assessed on simulated and real regression tasks.
\end{abstract}

\clearpage

\section{Introduction}

Suppose you are a terrible golfer invited to play a twisted form of golf. Rules are as follows. You are allowed 20 strokes. Performance is evaluated by a distance between a reported position and the hole. Two radically different strategies are available to the dismal golfer. The first is to try a hole in one 20 times and report the average position of the ball. By unbiasedness (despite high variance), the average shot should be close to the hole. A less flamboyant strategy is to do 20 small yet precise strokes (perhaps with the putter) and report the final position of the ball. The hole is the conditional mean, the golfer is a greedy algorithm, the first strategy is in the spirit of Random Forest (RF, \citealt{breiman2001}), while the second is based on Boosted Trees (BT, \citealt{freund1996experiments,friedman2001}).

This paper proposes a unified view of BT and RF by adopting a simple taxonomy. The two algorithms differ in two main aspects: model structure (additive shallow trees vs one deep tree) and how they tame the imperfect greedy optimization path (small learning rate vs randomization/diversification).  The first means that the space of potential predictive functions depends on the underlying model-building algorithm (e.g., additive vs multiplicative). The second is picking a strategy to successfully explore that space. 


\begin{table*} 
\caption{A Tree Ensemble Quaternity}\label{quad}
\vspace{-0.75cm}
\hspace{-0.75cm}
\setstretch{1.5}
\setlength{%
\arrayrulewidth}{0.99pt}
\par
\begin{center}
\begin{tabular}{l|l|c|c|}
\multicolumn{2}{c}{}&\multicolumn{2}{c}{\textbf{Model Structure}}\\
\hhline{~|~|-|-|}
\multicolumn{2}{c|}{}& \cellcolor{PineGreen!15}Additive Shallow Trees & \cellcolor{PineGreen!15}One Deep Tree \\
\hhline{~|-|-|-|}
\multirow{2}{*}{\textbf{Regularizer}}& \cellcolor{PineGreen!15}Slow Learning & Boosting & Slow Growing Tree \\
\hhline{~|-|-|-|}
&  \cellcolor{PineGreen!15}Randomization + Averaging & Booging & Random Forest \\
\hhline{~|-|-|-|}
\end{tabular}
\end{center}
 \vspace*{-0.35cm}  
\end{table*}

The base tree models contrasted in this paper are the one deep tree  and shallow additive trees. The first refer to fitting a single tree until there is little or nothing left to partition, making depth dependent on the sample size $N$. The second refer to sequentially (and additively) combining many trees of limited \textit{and} fixed depth, usually from 1 ("stumps") to 10. 

Model/regularization pairs are well known. BT fits an additive model and uses slow learning via its small learning rate $\nu$. RF performs fast but randomized learning of one deep tree, and gets the conditional mean right on average. To re-emphasize, the model-building path of additive shallow trees is traditionally regularized in one way, and that of one deep tree in another. Yet, there is no ex-ante reason for these strict associations: additive shallow trees could well be regularized by randomization/diversification and the one deep tree's variance could be reduced by constructing it slowly. While the former was already proposed in \citet{MSoRF} for other purposes, the latter remained until now a conceptual curiosity. This paper makes it operational by developing the missing corner in Table \ref{quad}: a Slow-Growing Tree (SGT). 

\begin{wrapfigure}{r}{0.5\textwidth} 
\begin{center} 
\vspace*{-0.25cm}  
\hspace*{-0.55cm}\includegraphics[width=.5\textwidth]{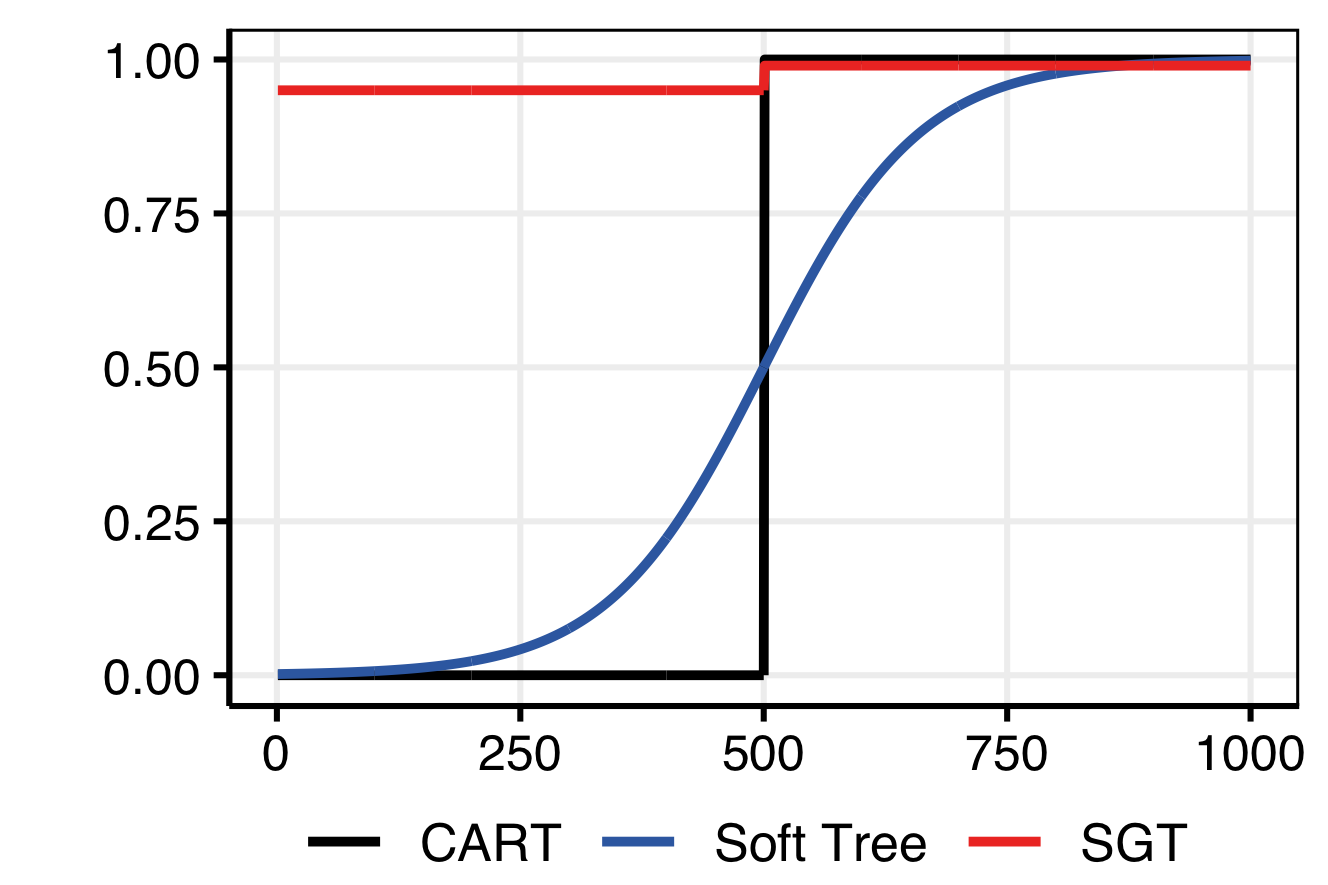}
\vspace*{-0.05cm}  
\caption{\footnotesize SGT's splitting rule ($\eta=0.05$) vs others.}
\label{sgt_vs_st}
\vspace*{-0.35cm}  
\end{center}
\end{wrapfigure}

SGT works by recognizing that CART \citep{breiman1984classification} can be seen as an extreme case of an iterative weighted least square procedure. More precisely, when splitting a sample into two subsamples, it gives a weight of one to observations that satisfy the splitting rule, and 0 to others. Then, calculations determining the next split for that particular node are carried using  observations with a weight of 1.  SGT introduces a learning rate $\eta$ such that observations not satisfying the splitting rule are allocated a weight of $1- \eta$ rather than 0.  In Figure \ref{sgt_vs_st},  SGT's splitting rule is contrasted with that of CART and Soft Trees \citep{jordan1994hierarchical,irsoy2012soft}. Thus,  unlike the two included alternatives,  whenever $\eta<1$,  observations at the far left are not completely discarded -- or at least, not by a single sweep of the greedy algorithm.\footnote{In fact,  Soft Trees as proposed in \cite{irsoy2012soft} are estimable by gradient descent, which comes with benefits \textit{and} costs.  More on this in section \ref{sec:related}.} Trivially,  SGT collapses to CART for $\eta=1$. Furthermore, it is easy to see how naturally shrinkage operates. A decreasing learning rate means that the children nodes' current average will be shrunk to its parent node's --- by allowing the other children's observations to enter subsequent calculations with a non-negligible weight.





This paper's contributions are threefold. \textbf{First}, SGT helps in showing how two famous algorithms (BT and RF) are linked through a tree taxonomy. A greedy algorithm tamed by randomization (RF) or slow learning (SGT) gives nearly identical results when they estimate the \textit{same model} (the one deep tree). Without SGT, we cannot hope to empirically demonstrate this equivalence because yes, BT uses slow learning, but it also fits a different model (additive shallow trees). By introducing SGT and Booging, which fill the missing corners of the tree ensemble quaternity, it is possible to distinguish the benefits of different regularizer vs different models. 

\textbf{Second}, SGT helps us understand RF, which frequent great performance is still not fully understood. RF tame the imperfect optimization of a one deep tree problem with randomization (bagging + model perturbation). The introduction of SGT shows that randomization in itself is \textit{not} the key, but it is only \textit{one} way of regularizing the optimization path of one deep tree. Further, SGT is shown to behave like  $l_1$ penalized kernel regression. This understanding of RF "by pieces" is useful for future model development as it shows that a successful "one deep tree" does not necessitate Bagging or feature randomization, which until now were indissociable. 

\textbf{Third,} SGT is interesting in itself by potentially providing gains on small data sets. When faced with 200 observations,  RF inevitably relies on trees of limited depth, giving a certain importance to early splits, which can be of disputable quality \cite{ESL}. In such situations, better results may be obtained from slow learning --- the choice of regularizer mattering more for small $N$. A prominent example is macroeconomic forecasting, where (pertinent) recorded history starts in the 1960s and is (at best) at the monthly frequency.\footnote{Using trees on time series data can seem odd at first glance. It is fine with \textit{stationary} data, which is true after applying standard econometric transformations. A new literature shows that tree ensembles are proficient at macro forecasting \citep{chen2019off,medeiros2019,GCLSS2018}. }







The new algorithm is evaluated against standard alternatives on simulated and real data sets. SGT behaves very similarly to RF, with sometimes marginally better/worse performance, supporting a \textbf{slo}w \textbf{t}ree \textbf{h}ypothesis (SLOTH) to explain RF.. That is, under a tree data-generating process,  SGT and RF coincides, and SGT can be seen as computationally feasible version of an extremely high-dimensional Lasso where regressors are interacted dummies constituting potential tree structures.  More generally,  this observation  is in line with the conjecture that the two regularization strategies (diversification vs slow learning) are largely interchangeable, and the best choice will depend on the data set. Hence, the development of SGT is conceptually interesting \textit{per se} since it permits to single out the effect of different regularization strategies on the \textit{same model} -- as highlighted in Table \ref{quad}.

Empirically backing the SLOTH, it is found that when the true DGP is a tree, SGT, and RF performances match. This observation helps link back RF to arguments that certain boosting procedures with infinitesimal learning rates approximate the Lasso path \citep{ESL,rosset2004boosting} -- i.e., a global solution to a high-dimensional problem. When it comes to SGT (and RF), the high-dimensional space of generated regressors just got wider, and which corners of it that get to be explored depend on the previous greedy steps. Finally, Booging (bagging stochastic gradient boosted trees with a high learning rate)  can deliver gains over BT, challenging slow-learning as the de facto regularization choice for additive trees. 

This paper is organized as follows. In section \ref{sec:sgt}, I detail the main conceptual insights, present the SGT algorithm and discuss practical aspects. Sections \ref{sec:simuls} and Section \ref{sec:empirics} report results on simulated and real data sets, respectively. Section \ref{sec:con} concludes.



\section{Ensembling Trees Revisited}\label{sec:sgt}




BT is a greedy algorithm sequentially combining trees to minimize a loss function. The resulting prediction is
$$\hat{y}_i = \sum_{s=1}^S \nu \mathcal{T}_s(X_i)$$
where $y_i \in {\rm I\!R}$ is a target variable, $X_i \in {\rm I\!R}^K$ are predictors, $\nu$ is a learning rate, and $S$, the number of steps (and trees), is a tuning parameter governing early stopping. Intuitively, its range of candidate values depends directly on $\nu$ -- smaller values of $\nu$'s typically necessitate higher values of $S$. In practice, it is customary to tune both \citep{friedman2002}. Very often, the lower $\nu$ gets, the better generalization is \citep{friedman2001}. 


Inspired by the above, I propose an algorithm estimating (slowly) a \textit{single} tree that can match (and sometimes beat) RF's performance.  While the utilization of a learning rate comes very naturally in a Boosting context (or in a deep learning one, see \citealt{deeplearning}), it is less clear how it should materialize when fitting a single deep tree. A useful observation is that CART is an extreme form of an iterative weighted least squares (WLS) procedure. The proposed implementation leverages the WLS analogy to develop a less "radical" tree-building algorithm.



\nocite{buhlmann2003boosting}
\nocite{buhlmann2007boosting}
\nocite{freund1996experiments}
\nocite{mason2000boosting}

Consider the deceptively simple tree below, which is obtained after one recursion of CART.

\vspace{0.25cm}
\Tree[.{Full Sample} 
[.{$A \equiv \{i | x_i < 0\}$} 
 ]
[.{$A^C \equiv \{i | x_i \geq 0\}$} ]
 ]
\vspace{0.25cm}

The subsequent problem to grow the tree on the $A$ side is
\begin{equation}
\begin{aligned}
(k^*_A,c^*_A)=\argmin\limits_{k\in \{1,...,K\}, \smallskip c \in {\rm I\!R}}\Bigg[ &\min\limits_{\mu_{1}} \sum \limits_{\{ i | X_{i}^k \leq c  \}} \omega_{i}^A \left(y_{i}-\mu_{1}\right)^{2}  + &\min\limits_{\mu_{2}} \sum \limits_{\{ i | X_{i}^k > c  \}} \omega_{i}^A \left(y_{i}-\mu_{2}\right)^{2}\Bigg]
\end{aligned}
\end{equation}
where $\omega_i^A = I(i \in A)$. This can be generalized to $$\omega_i^A = I(i \in A)+(1-\eta)I(i \in A^C)$$ where $\eta \in (0,1]$ is a learning rate. Trivially, those new weights collapse to those of CART when $\eta=1$. Like $\nu$ in Boosting, $\eta<1$ shrinks the new problem to the old one. This partial rejection of the greedy algorithm's proposed split desirably moderates the chosen optimization direction. This can also be seen as softly pruning the tree at each step $s$: with a decreasing $\eta$ the mean in the current children node is shrunk back to full sample mean, partly annulling the earlier split. Of course, the idea here, like in Boosting, is that there will be many more such $s$'s than when $\eta=1$, allowing SGT to build a deeper tree where each specific "split" matters less on its own. 


\begin{algorithm*}[tb]
   \caption{Slow-Growing Tree}
   \label{sgt_algo}
\begin{algorithmic}
   \STATE {\bfseries Input:} Training data $\left[y_i \enskip \boldsymbol{X}_i\right]$, test set predictors $\boldsymbol{X}_j$, learning rate $\eta \in (0,1]$, maximal concentration index $\bar{H}$
   \STATE Initialize $\omega_i^{0}=1 \enskip \forall i$.
   \FOR{$l$'s such that $H_l <\bar{H}$}
   \STATE \hspace{0.25em} $(k^*_l,c^*_l)= \argmin\limits_{k\in \{1,\dots,K\}, \smallskip c \in {\rm I\!R}}\left[\min\limits_{\mu_{1}} \sum \limits_{\{ i | X_{i}^k \leq c  \}} \omega_{i}^l \left(y_{i}-\mu_{1}\right)^{2}+\min\limits_{\mu_{2}} \sum \limits_{\{ i | X_{i}^k > c  \}} \omega_{i}^l \left(y_{i}-\mu_{2}\right)^{2}\right]$
   \STATE Create 2 children nodes, one with $\omega_{i}^{l} \leftarrow \omega_{i}^{l} (1-\eta I (X_{i}^{k^*_l} \leq {c^*_l} ))$, the other with $\omega_{i}^{l} \leftarrow \omega_{i}^{l} (1-\eta I (X_{i}^{k^*_l} > {c^*_l} ))$.
   \ENDFOR
   \STATE {\bfseries Return:} $\hat{y}_{j}=\sum_{l=1}^L \left( w_j^l(\boldsymbol{X}_j) \sum_{i=1}^N \omega_i^{l}y_i \right)$ where $w_j^l(\boldsymbol{X}_j) = \frac{\omega_j^{l}(\boldsymbol{X}_{j})}{\sum_{l=1}^L \omega_j^{l}(\boldsymbol{X}_{j})}$
\end{algorithmic}
\end{algorithm*}
	
Algorithm \ref{sgt_algo} outlines SGT's recursive procedure. A few remarks are in order. First, a typical stopping rule for CART is either the minimal size of the node before attempting a last split, or a minimal size of the terminal nodes. Anyway this is set up, it is customary that tree constituents in RF are approximately fully grown with leaves containing between 1 and 5 observations. By construction, all nodes of SGT, no matter how deep within the tree they appear, all incorporate $N$ observations. The barometer of in-sample learning is  an increasingly unequal distribution of $\omega_i^l$'s, where $\omega_i^l$ is the weight of observation $i$ in a given leaf $l$. Thus, the stopping rule shall be formulated in terms of the imbalance of $\omega_i^l$. I use an upper bound $\bar{H}$ on the Herfindhal index $H_l \equiv \sum_{i=1}^N (\omega_i^{l})^2$, which is a widely used measure of industry \citep{DOJ}, labor market \citep{azar2020labor}, and portfolio \citep{avila2013concentration} concentration. That bound is set between 0.2 or 0.3 in practice, which is extremely high (that is, the distribution is highly unequal with only a few observations having a noticeable weight) and is thought of approximating what a fully grown (overfitting) tree would be in this new environment. More precisely, in a standard tree, if a terminal node includes 5 observations, then its $H_l=5\times (\sfrac{1}{5})^2+(N-5)\times 0^2=0.2$. Considering smaller $\bar{H}$'s (like 0.05 or 0.1) is a form of early stopping which will be shown to pay off in low signal-to-noise environments in section \ref{sec:simuls}. 

Weights for a given terminal node is the product of all filters applied to get from the top of the tree to that particular leaf. This is, again, a generalization of the usual tree weights where filters (splitting rules) have the additional property of setting part of the weights directly to 0. In that spirit, the mean of that terminal node is a weighted average with weights $\omega_i^{l}$. Finally, since the hard separation embedded in CART (observations go on to $A$ or $B$ but not both) is no more, a new data point $j$ will almost surely fall in more than one leaf (depending on its $\boldsymbol{X}_j$). Those within-leaf weighted averages ($\sum_{i=1}^N \omega_i^{l}y_i$) must be combined to get the final prediction. Taking naively the simple average would fail as leaves where $j$ has a very small $\omega_j^l$ would get the same preponderance as those where $\omega_j^l$ dominates. $w_j^l(\boldsymbol{X}_j) = {\omega_j^{l}(\boldsymbol{X}_{j})}/{\sum_{l=1}^L \omega_j^{l}(\boldsymbol{X}_{j})}$ adjusts for that. It is easy to see that is the $\eta=1$ case where weights are dichotomous (and thus leaves do not overlap) that $w_j^l(\boldsymbol{X}_j)=1$ for a single leaf and zero for all others.

Lastly, a very technical, yet necessary, consideration. The fact that a split can be invoked twice (or as often as needed, really) gives rise to a particular situation that would not arise in CART: the distribution of $\omega_i$ in a node can revert back to the initial (flat) one. For instance, if we "split" using $x_i < 0$ in a node which unique previous split was based on $x_i < 0$, we get two new children nodes. The first is not problematic -- we simply get twice the discriminating action of $x_i < 0$, that is, a weight of 1 for observations that satisfy the rule, and $(1-\eta)^2$ for those that do not. However, the second child node will have a flat distribution of weights because some observations got down weighted once by $(1-\eta)$ in the first split, and all the remaining children get down weighted once $(1-\eta)$ by the second split. Hence, they all have a weight of $(1-\eta)$, which, after normalization, are the initialization weights. If left unattended, this creates an infinite loop of a tree within a tree. The solution is simple: such dead branches are trimmed out. Note that this is not problematic in SGT as the assigned observations are appearing elsewhere anyway (in CART, this would create a hole in the predicting function).
	
\subsection{What is SGT estimating?}

I start by reviewing a popular interpretation of what a low learning rate does to a linear boosting (or forward stagewise regression) problem. Then, I leverage that knowledge to propose an explanation of what SGT is estimating    

\subsubsection{Boosting as a $l_1$-regularized Problem}

\citet{ESL} observe that a simpler form of boosting (without trees and observations reweighting) delivers coefficients paths indistinguishable to that of Lasso \citep{tibshirani1996}. More precisely, provided a small enough learning rate, plotting coefficients as functions of decreasing regularization strength -- i.e., $\hat{\beta}_{\text{Boosting}}(s)$ against an increasing $s$ and $\hat{\beta}_\text{Lasso}(\lambda)$ for a decreasing $\lambda$ -- gives rise to identical figures. This relationship is further explored in \citet{rosset2004boosting} where boosting is argued to be a computationally convenient approach to solve a (very) high-dimensional regression with $l_1$ regularization.\footnote{Closely related, \citet{zhang2005boosting} provide a consistency result for Boosting provided a small step size and early stopping.} This exact equivalence hinges on $\hat{\beta}_\text{Lasso}(\lambda)$ paths being monotone since Boosting solves a monotone Lasso problem where coefficients are penalized according to their "arc-length" \citep{hastie2007forward}.\footnote{Unlike traditional Lasso, the monotone version cannot be cast as a global optimization problem. Nonetheless, monotone Lasso has a less wiggly regularization path and can provide gains over its famous brother in high-dimensional environments where regressors are highly collinear \citep{hastie2007forward}.}


\citet{schapire2013explaining} nuances those claims by pointing out that the  $l_1$ argument applies to a very specific variant of Adaboost, and that the Lasso analogy is hard to reconcile with the frequent observation that a very large $S$ will not substantially damage out-of-sample performance -- which would certainly not happen for Lasso in a classic setup when letting $\lambda\rightarrow0$. On the other hand, \citep{kobak2020optimal} show that the optimal $\lambda$ (for ridge) can be either negative or 0 in extremely high dimensions -- which is where BT operates. This provides an avenue to conciliate the $l_1$ "explanation" of boosting and the empirical observation that the generalization sometimes does not decrease in $s$.\footnote{This is closely related to modern "interpolating regime" or "double descent" arguments  \citep{wyner2017explaining,belkin2019reconciling,bartlett2020benign}.}




\subsubsection{SGT Estimates the True Latent Tree}

Like CART, SGT minimizes the empirical loss by creating more and more bins that are increasingly discriminatory. The key difference is that SGT’s path towards highly polarized bins is constituted of many splits that are only mildly potent by themselves.  But there is more. As a last building block behind the SLOTH,  SGT can be cast as a peculiar Boosting problem. Then, we can exploit \citet{rosset2004boosting} and \citep{hastie2007forward}'s arguments that Boosting, under certain conditions, performs a high-dimensional Lasso -- i.e., solving a \textit{global} problem.   

A symmetric 2-layers tree's prediction can be written as
\begin{equation}\label{tree}
\begin{aligned}
\hat{y}_i = \enskip & I(x_{i}>0)\left[\alpha_1 I(z_{i}>0)+\alpha_2 I(z_{i}\leq 0)\right]  
\enskip +& I(x_{i}\leq 0)\left[\gamma_1 I(q_{i}>0)+\gamma_2 I(q_{i}\leq 0)\right]
\end{aligned}
\end{equation}
where $\gamma$'s and $\alpha$'s are within-node mean parameters. $x$, $z$ and $q$ are predictors. Define $d_{x,i}^{+}=I(x_{i}>0)$ as a regressor and the rest accordingly. Note that SGT can be obtained by letting $d_{x,i}^{+}=\tilde{\eta} I(x_{i}>0)+(1-\tilde{\eta})I(x_{i}\leq 0)$ where $\tilde{\eta}$ is some normalized version of $\eta$. \eqref{tree} is also an additive model
\begin{equation}\label{polytree}
\begin{aligned}
\hat{y}_i = \theta_1 d_{x,i}^{+} d_{z,i}^{+}+\theta_2  d_{x,i}^{+} d_{z,i}^{-} +\theta_3 d_{x,i}^{-}d_{q,i}^{+}+\theta_4 d_{x,i}^{-} d_{q,i}^{-}.
\end{aligned}
\end{equation}
with "polynomials" of order 2. A tree with 3-layers would extend \eqref{polytree} to 8 polynomials of order 3.  Finally, one can rewrite \eqref{polytree} using the identity $d_{z,i}^{+}=1-d_{z,i}^{-}$ (and similarly for $q$) to obtain
\begin{equation}\label{polytree2}
\begin{aligned}
\hat{y}_i &=  d_{x,i}^{+}\left[(\alpha_1-\alpha_2) d_{z,i}^{+}+\alpha_2\right] 
+ d_{x,i}^{-}\left[(\gamma_1-\gamma_2) d_{q,i}^{+}+  \gamma_2\right] \\
&= \beta_1 d_{x,i}^{+}+\beta_2  d_{x,i}^{+} d_{z,i}^{+} +\beta_3 d_{x,i}^{-}+\beta_4 d_{x,i}^{-} d_{q,i}^{+}.
\end{aligned}
\end{equation}
As in \eqref{polytree}, \eqref{polytree2} is additive but now, it is explicit that "earlier" basis functions are not ejected from the model. Hence, the deep tree can also be cast within \citet{rosset2004boosting} and \citet{hastie2007forward}'s $\hat{y}_i=h(\boldsymbol{X}_i)'\beta$ form where $h\in\mathcal{H}$. $\mathcal{H}$ is the fixed space of \textit{all possible basis expansions} for the one deep tree, and it is significantly larger than what additive shallow trees can allow for. The exploration of $\mathcal{H}$ is made possible by relying on the past steps of the greedy algorithm to guide its search through higher-order polynomials. That is, as an implication of the tree structure, some part of "later" basis expansions is already fixed.  

Using this representation, the intuition provided in \citet{rosset2004boosting} directly applies, provided $\beta$'s are monotone. This can be obtained by designing new basis functions using a rule which follows from a series of casual observations. First, it is noticed that one sweep of SGT's greedy algorithm updates two $\beta$'s --- that of the older basis expansion (like $d_{x,i}^{+}$) and that of the newly added one ($d_{x,i}^{+} d_{z,i}^{+}$). Second, it does not matter for the span of the model whether we expand with $d_{x,i}^{+} d_{z,i}^{+}$ or $d_{x,i}^{+} d_{z,i}^{-}$.  Third, given that those specific basis expansions amount to splitting a mean into two means, by construction, one of them will always be at least as high (in absolute value) as the parent node's mean. Thus, we can choose to use $d_{x,i}^{+} d_{z,i}^{+}$ or $d_{x,i}^{+} d_{z,i}^{-}$ so that the updated coefficient on $d_{x,i}^{+}$ (which is already in the active set of regressors) satisfies monotonicity. 

As a result of the above, SGT is a high-dimensional $l_1$ regularized regression with an extremely large $\mathcal{H}$ which is explored at each $s$ in a particular way. Naturally, that space is the space of trees of a certain dimension (bounded by the size of the data set). SGT is a computationally cheaper way of estimating the true latent tree, whose direct global optimization would be an extremely high-dimensional Lasso problem. Moreover, $l_1$ regularization in SGT's context can be straightforwardly interpreted as pruning: shrinking $\beta_2$ completely to 0 prunes the  $x_i>0$ side of the tree. Selecting both $\beta_2$ and $\beta_4$ to 0 prunes the 2-layers tree a single layer one. Additionally, early stopping takes a similar meaning. 

Since RF is yet another way of taming CART, this understanding of SGT is transferable to RF.  In \citet{MSoRF}, it is argued, by an analogy to forecasting with nonlinear (recursive) time series models, that bagging brings an imperfect greedy procedure to extract the true conditional mean.  The deeper reason is that taking an expectation over a series of \textit{nonlinear} recursions gives a very different outcome from that of a series of recursions that take expectations as inputs. The former is what we hope CART would do, and the latter is what it actually does. Bagging allows simulating the right expectation \citep{MSoRF}. The SLOTH is that this performs the \textit{same} role as what we expect from the learning rate. A conjecture as to why a low $\eta$ deliver similar results is that, by shrinking the children weights to those of the parents, this pushes the "updating" mapping towards the identity mapping, which is \textit{linear}, making the expectation in recursion problem much milder.  Simulations will back the SLOTH  by showing that when the true DGP is a tree, SGT and RF provide near-identical performance.








\subsection{Interpretation?}


In RF, the visual inspection of underlying trees is doomed because there are usually 500 of them and they are all of substantial size. The former problem remains for BT, whereas the latter vanishes since trees are typically shallow.  However, unlike RF, all those trees are fitted to a different target (the pseudo-residuals at step $s$), a new complication. Thus, it is customary to attempt the understanding of such models using variable importance measure, partial dependence plots and surrogate models \citep{molnar2019interpretable}. Yet, none of those options provide the enviable clarity of inspecting a single tree. 

SGT provides a single tree. However, the utilization of the learning rate means giving up hard-thresolding, which implies that a single observation can belong at varying degrees to different leaves. Also, $\eta<1$ allows for splitting rules to be invoked more than once. This gives rise to "hierarchy" of splits: some on which the greedy algorithm really insists, and those that it may have merely stumbled upon along the way. None of those complications prevents from plotting the corresponding tree. For instance, the fact that some observations are shared across leaves can be simply thought as tree obtained on an artificially augmented data set ($N$-wise). Rather, the core roadblock is that the resulting SGT is very large, making human inspection daunting. In sum, SGT opens new avenues for the interpretation of a highly performing model (matching in many ways RF), but exploring those is non-trivial and is kept for future work.

\subsection{Hyperparameters and Computational Aspects}

Like any tree-based model, computational demand increases in $N$ and $K$. It depends highly on $\eta$: a smaller $\eta$ necessarily increase the depth of SGTs, the same way $\nu$ necessitate a higher $S$. Hence, computational burden behaves similarly to that of fitting a deep tree on a very large data set ($N$-wise). Additionally, since we are recursively solving WLS problems rather than assigning 0-1 weights (à la CART), the size of samples used to pick $(k^*_l,c^*_l)$ at the lowest layers of the SGT is the same as those of upper layers. Finally, with a lower $\bar{H}$ acting as an early-stopping rule, computations decrease in $\bar{H}$. To accelerate computations -- which can be demanding for large $N$ data sets, I let $\eta$ increase marginally with depth so the algorithm avoid languishing with many leaves with a $H_l$ slightly below $\bar{H}$. The rule of thumb is starting at $\eta=0.1$ and increasing it by 0.01 with depth (that is, how many splits led to the current node) until an imposed plateau of 0.5 is reached. To further increase speed and add mild regularization à la stochastic gradient descent \citep{friedman2002}, $\texttt{mtry}$ =0.75 is used throughout. 

The bias-variance trade-off within SGT operates like in Boosting. Deeper trees (as controlled by $\bar{H}$) will decrease bias at the expense of variance (like adding more trees in BT). Also, as it is often found for BT (and originally documented in \citealt{friedman2001},) the best learning rate for SGT is almost always the smallest one and there is little benefit in early-stopping. 


\subsection{Related Works}\label{sec:related}

Related to SGTs are soft trees \citep{irsoy2012soft}. 
They differ from SGT in many ways -- goals and results. They consist in changing the hard-threshold rule of CART by a sigmoid function which will still allocate 0 and 1 weights to observations in the tails, but will have less drastic weight change around $c^*$. Hence, some observations will still be completely discarded  on the basis of a single split on a single variable, and as a result, soft trees still suffer from most of the CART ailments -- except that of not attaining a global optimum since it can be estimated by gradient descent. The "optimality" concern has been the subject of recent work  \citep{bertsimas2017optimal,hu2019optimal,norouzi2015efficient,blanquero2020sparse} where recursive partitioning is retired in favor of global optimization techniques. While those methods will almost always beat CART, their performance usually lags behind that of RF in a sizable way (see \citet{bertsimas2017optimal} results on classification and those of \citet{blanquero2020sparse} for regression).

SGT's relationship with Soft Trees can benefit from further discussion. Despite appearances, SGT has  little to do with Soft Trees in motivation, implementation, and outcomes. That also applies to fuzzy trees or probabilistic versions of CART where smoothness also emerges. Figure \ref{sgt_vs_st} showcases the difference in splitting rules. A single SGT sweep does \textit{not} assign probabilities to the nodes as Soft Trees would. They key element is rather the \textit{learning rate} which shrinks the orange line (CART) to a flat line at 1 (no learning).  The blue and black lines separate observations 1 to 30 and 70 to 100, and there is no turning back. As is clear from Figure \ref{sgt_vs_st}, the red one does \textit{not} separate "forever" and rather use the greedy algorithm as suggestive of how weights should be adjusted. As a result, many SGT iterations are necessary for the data to be really split.  Of course, this is the "one deep tree" version of Boosted Trees (BT) adding the next best tree (found greedily) but shrinking it severely to zero. 

If there were a single regressor to split on, and the true split is smooth and with respect to $X$, then we would expect SGT to reproduce the soft tree split as a sum of many slow splits in different positions along the x-axis. RF could do the same (see \cite{buhlmann2002analyzing}) by relying on averaging fast-learning rules. But these are very special circumstances. In an environment where different regressors could be used for the split and the smoothness of the split is unknown, SGT and Soft Trees will radically differ. An advantage of SGT and RF over Soft Trees and CART is that by construction, splits need not be made of a single variable (both SGT and RF are "ensembling" in their own way) while Soft Trees still make  large learning leaps by fully splitting on a \textit{single variable}. 

A more general question emerges. Given the popularity of neural networks, is there any use for another greedy algorithm -- which does not reap the well-document benefits of gradient descent? The answer is yes. While Soft Trees and their variant may be estimable by gradient descent, some more subtle advantages of greedy algorithms (especially for small tabular data) are lost in the process. First, automatic feature selection (which RF and BT excel at) is often ejected from the process \cite{arik2019tabnet}, a globally-optimized model needing to be globally specified first.\footnote{See \cite{arik2019tabnet} and references therein on how to partly generate automatic feature selection in deep networks.} In contrast, in greedy methods, model building and estimation happen simultaneously. Second, greedy methods appear more apt at dealing with noisy labels. \cite{MSoRF} makes that point that RF performs implicit optimal early stopping out-of-sample and \cite{wyner2017explaining} discuss potential sources of the incredible robustness of BT. As a result of those observations (and likely others), BT reigns over many Kaggle competitions involving tabular data, and RF is becoming an inevitable benchmark for economic/financial data \cite{MSoRF}.





\subsection{\textit{Booging}}


Booging (for Bagging Boosting) is the application diversification-based regularization to additive shallow tree structure. It was originally proposed in \citet{MSoRF} to show that any randomized greedy algorithm (beyond RF) can be self-regularized, that is, perform implicit early stopping. In this paper's simulations and empirical work, it will be used again to show how Booging can rival BT with tuned learning rates -- further demonstrating the virtues of the championed tree ensemble quaternity.  

Booging consists of running BT with a relatively high learning rate (like $\nu=0.25$) on many bootstrap samples. Clearly, Bagging is only useful here because each BT run becomes increasingly unstable with high $\nu$'s. Moreover, each run is perturbed by using Stochastic Gradient Boosting \citet{friedman2002} with a certain fraction (often 0.3 or 0.5) of randomly selected observations being used to construct trees. To reinforce diversification potential (especially when regressors are scarce), the Booging also uses data augmentation (i.e., include noisy carbon copies of $\boldsymbol{X}$). For all implementation details, see \citet{MSoRF}. 


\section{Simulations}\label{sec:simuls}

I use simulated data from five standard data generating processes (DGPs) to evaluate when SGTs and Booging outperforms their counterparts relying on their traditionally  associated regularization scheme.

\begin{figure*}[tb] 
\begin{center} 
\hspace*{-0.2cm}\includegraphics[scale=.58]{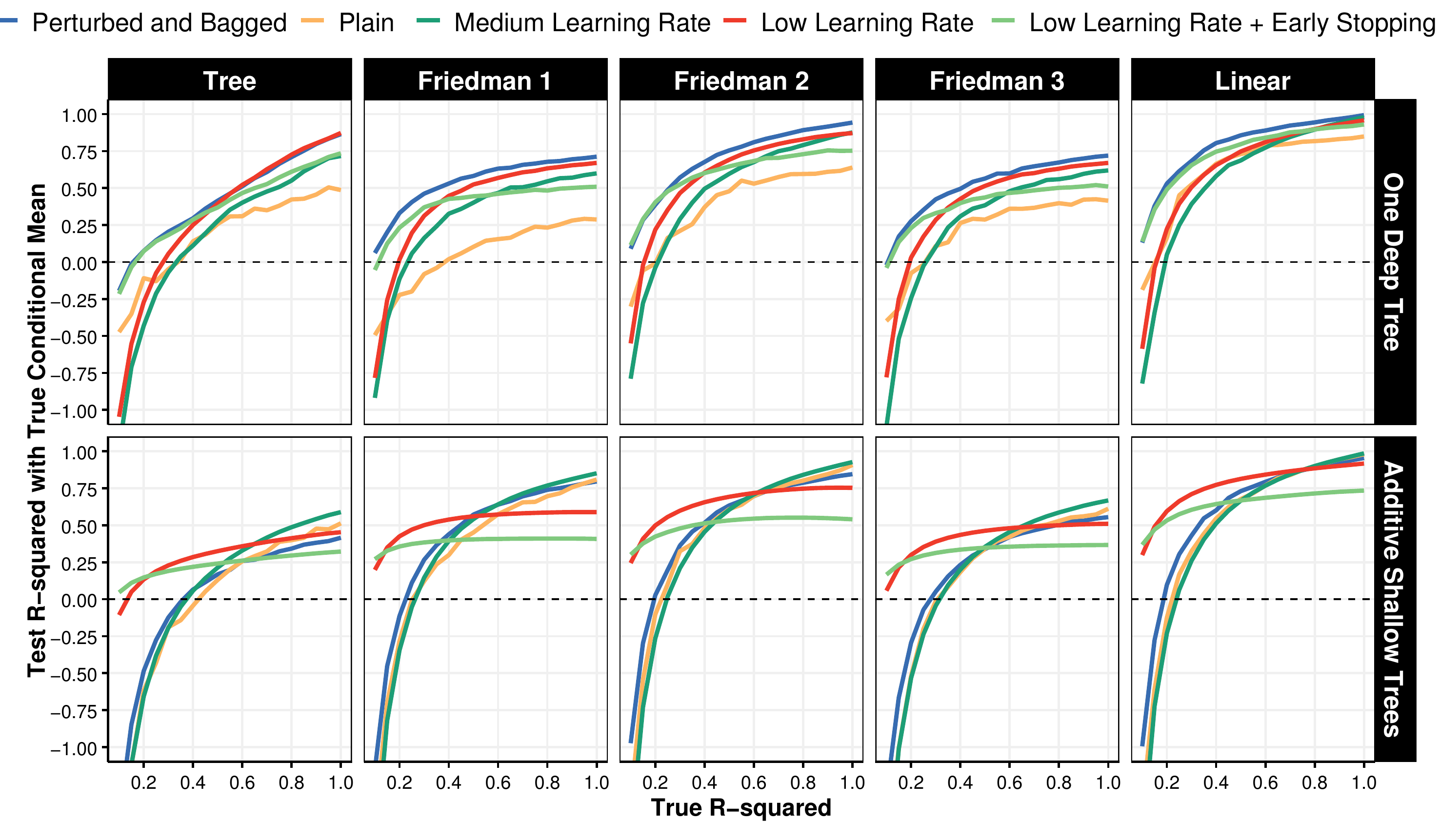}
\vspace*{-0.3cm}  
\caption{\footnotesize This plots the hold-out sample $R^2$ between the prediction \textbf{and the true conditional mean}. The level of noise is decreasing along the $x$-axis. Column facets are DGPs and row facets are "models". The $y$-axis is cut at -1 to favor readability because a few models go largely below it for the lowest signal-to-noise ratio case. Note that negative $R^2$'s mean the evaluated model does worse than the sample mean \textit{on the test set}.}
\label{simul_1}
\end{center}
 \vspace*{-0.65cm}  
\end{figure*}

Five versions of the two classic tree models (one deep tree, additive shallow trees) are considered. In Figure \ref{simul_1}, each model is a row facet. Both are constructed in five different ways, each with their own associated color. First, we have the bagged and perturbed version, which corresponds to RF for one deep tree, and Booging for additive shallow trees. For RF, both \texttt{mtry} and the minimal node size are tuned by cross-validation. The "plain" versions of the two structures are CART (with a tuned depth) and BT with a high $\nu$ (0.25) and a tuned number of trees. Third, I consider 3 versions with varying learning rates with one of them additionally incorporating early stopping. For the "one deep tree", these are SGTs with $\eta\in \{0.5,0.1,0.1\}$ and  $\bar{H} \in \{0.25,0.25,0.05\}$. For additive trees, this is standard BT with $\nu \in \{0.1,0.001,0.001\}$ and $S\in \{1500,1500,750\}$.\footnote{As for other standard hyperparameters of Boosting-based models, the fraction of randomly selected observations to build trees at each step is 0.5, and the interaction depth of those trees is 5. $\nu$ is set to 0.25 for Booging.} 

The five DGPs are identical to that of \citet{MSoRF}.  First, we have the "Tree" DGP which is generated using a CART algorithm's prediction function as a "new" conditional mean function from which to simulate. The depth of the tree is set by fixing the minimal node size so that true terminal nodes contain about $\sfrac{1}{8}$ of the sample on average. Then, we have Friedman 1, 2 and 3 \citep{MARS} as well as a linear model. The latter's conditional mean is the sum of five mutually orthogonal and normally distributed regressors. Each model is fed 10 predictors with the fraction of truly useful ones varying across DGPs (but is usually around 50\%). In Figure \ref{simul_1}, DGPs are column facets.

Finally, it is expected that the performance differential between the two regularization schemes depends not only on the DGP but also the signal-to-noise ratio. To explore such dependence, the performance is reported for "true" $R^2$'s ranging (on Figure \ref{simul_1}'s $x$-axis) between 0.1 and 0.99.     For all simulations, $N=100$ and the test set also has 100 observations.\footnote{Relative performance results are close to identical when bumping $N$ to 400.}  The reported performance statistic is the mean $R^2$ between hold-out sample predictions and the true conditional mean for 30 simulations. Similarly to \cite{kellyml}, $R^2$ are used here and in the empirical section because (i) the focus is on regression, (ii) they deliver a ranking identical to that of RMSEs, (iii) they are in comparable (and intelligible) units from one simulation/data set to another, and (iv) carry pertinent information about the complexity of underlying task and the signal-to-noise ratio.

First, let us focus on the "Tree" column of Figure \ref{simul_1}. The blue line of RF is visually absent, because it is perfectly matched by SGT$\left(\eta=0.1, \bar{H}=0.05 \right)$ (light green) up until the true $R^2$ is about 0.45, and by SGT$\left(\eta=0.1, \bar{H}=0.25 \right)$ (red) afterwards.\footnote{It is understood that tuning $\bar{H}$ would simply generate a line close to the upper envelope of SGT$\left(\eta=0.1, \bar{H}=0.05 \right)$ and SGT$\left(\eta=0.1, \bar{H}=0.25 \right)$, and thus (tuned) RF.} This simply means that when it comes to estimating a conditional mean corresponding to a true tree, SGT and RF coincides -- as the SLOTH suggests. That is, slow learning and diversification are equally valid methods to tame CART's greedy algorithm.

In line with common wisdom for $\nu$ in BT \citep{friedman2001,rosset2004boosting}, the SGT$\left(\eta=0.1, \bar{H}=0.25 \right)$ line is strictly above that of $\left(\eta=0.5, \bar{H}=0.25 \right)$ (dark green). The strict dominance of a lower $\eta$ is also verified for the remaining four DGPs. Finally, CART (light orange) is noticeably inferior to any SGTs, provided $\bar{H}$ would be tuned appropriately. In fact, it is interesting to notice how parallel are the light green and orange lines for all DGPs, with the latter always below the former by a sizable constant margin.

For DGPs Friedman 1,2 and 3, RF has a small edge over SGTs which is mostly apparent for high signal-to-noise ratio environments -- RF performance is generally matched by SGT$\left(\eta=0.1, \bar{H}=0.05 \right)$ for noisier setups. In the case of the linear DGP, RF performance is almost exactly matched by SGT$\left(\eta=0.1, \bar{H}=0.05 \right)$ for all noise levels.

For the additive tree row, it is clearly visible that Booging implicitly performs early stopping, as the blue line closely follows the orange one, being marginally better or worse at times. Bringing $\nu$ down to 0.01 helps marginally, especially in the high true $R^2$ scenarios, whereas decreasing it even further to 0.001 pays off mostly when the true $R^2$ is low. As expected, the additive shallow trees, mostly underperforms with respect to the one deep tree when the true DGP is a tree. Another takeaway is that Booging, by using diversification with a high learning rate, is often comparable to a lower learning rate version of BT.

\section{Empirics}\label{sec:empirics}

I consider data sets that are standard in the literature (all from UCI repository or Kaggle) as well as 12 data sets representing different macroeconomic forecasting tasks.\footnote{For \textit{California Housing}, a subset of 2000 observations is used.} The latter contrast sharply with the traditionally considered regression examples:  it is time series data, the number of features is extremely high, the number of observations is low, and the signal-to-noise ratio is low when the data is properly transformed to induce stationarity. In short, it is a hostile environment where alternative regularization may yield much needed improvements.

The first 6 of those economic datasets are quarterly US data based on \citet{mccracken2020fred} and the popular \href{https://fred.stlouisfed.org/}{FRED} database from the Federal Reserve Bank of St-Louis. The 3 targeted macroeconomic variables are GDP growth, unemployment rate change (UR), and CPI inflation (INF), which are predicted at horizons of 1 quarter and 2 quarters. The predictors matrix is based on \citet{MRF} and \citet{MDTM}'s recommendations for ML algorithms applied on stationary (see \citet{mccracken2020fred} on that) macro data. Each data set has 212 observations (ending in 2014) and around 600 predictors.\footnote{The features set varies because of a mild screening rule.}

\begin{table} 
\begin{center}
  \begin{threeparttable}
\centering
\footnotesize
\caption{Test set $R^2$'s for all data sets and models \label{results1}}
\setlength{\tabcolsep}{0.5em}
\renewcommand{\arraystretch}{1.5}
  \rowcolors{2}{white}{gray!15}
\sisetup{detect-weight=true,detect-inline-weight=math}
\begin{tabular}{ l *{9}
S[
    table-format            = -2.3,
    input-close-uncertainty = ,
    input-open-uncertainty =  ,
    table-space-text-pre   = ( , 
    table-space-text-post  = \stars{***},
    table-align-text-post  = false
  ]{c} }
\toprule
\hspace*{0.5cm} &
\multicolumn{3}{c}{Benchmarks} &
\multicolumn{2}{c}{One Deep Tree} &
\multicolumn{2}{c}{Additive Shallow Trees} \\
\cmidrule(lr){2-4} \cmidrule(lr){5-6} \cmidrule(lr){7-8}
&
\makebox[3em]{AR} &
\makebox[3em]{Lasso} &
\makebox[3em]{Tree} &
\makebox[3em]{RF} &
\makebox[3em]{SGT} &
\makebox[3em]{BT} &
\makebox[3em]{Booging} \\
\midrule

Boston   &   & 0.67& 0.58& 0.88& 0.86***&\phantom{-} \textcolor{blue}{\bfseries  0.90}&0.88\tabularnewline
Abalone   &   & 0.51& 0.31&\phantom{-}  \textcolor{blue}{\bfseries  0.57}& 0.55**& 0.56&0.55\tabularnewline
Auto     & & 0.67& 0.55&\phantom{-}  \textcolor{blue}{\bfseries  0.71}& 0.65***& 0.68&0.66\tabularnewline
Red Wine  &   & 0.35& 0.25&\phantom{-}  \textcolor{blue}{\bfseries  0.47}& 0.42***& 0.41&0.37***\tabularnewline
Concrete    &   & 0.59& 0.41& 0.90& 0.86***&\phantom{-}  \textcolor{blue}{\bfseries  0.92}&0.90***\tabularnewline
Fish Toxicity    &   & 0.55& 0.36&\phantom{-}  \textcolor{blue}{\bfseries  0.64}& 0.64& 0.63&0.62\tabularnewline
NBA salary   &   & 0.49& 0.22&\phantom{-}  \textcolor{blue}{\bfseries  0.64}& 0.60& 0.54&0.52\tabularnewline
CA Housing    &   & 0.69& 0.51&\phantom{-}  \textcolor{blue}{\bfseries  0.81}& 0.77***& 0.78&0.80\tabularnewline
Ohio Housing   &   & 0.96& 0.79& 0.97& 0.96&\phantom{-}  \textcolor{blue}{\bfseries  0.98}&0.97\tabularnewline
Airfoil Noise   &   & 0.51& 0.36& 0.92& 0.85***&\phantom{-}  \textcolor{blue}{\bfseries  0.93}&0.87***\tabularnewline
\midrule
US-GDP-1Q & 0.27& 0.27& 0.09& 0.36& 0.34& 0.35&\phantom{-}  \textcolor{blue}{\bfseries 0.38}\tabularnewline
US-GDP-2Q&0.17&-0.02&-0.25& 0.17& 0.02**&\phantom{-}  \textcolor{blue}{\bfseries  0.19}&0.19\tabularnewline
US-UR-1Q &0.53& 0.45&-0.07& 0.58& 0.45**& 0.62&\phantom{-}  \textcolor{blue}{\bfseries 0.64}\tabularnewline
US-UR-2Q &0.29& 0.21&-0.25& 0.41& 0.36&\phantom{-}  \textcolor{blue}{\bfseries 0.48}&0.48\tabularnewline
US-INF-1Q &0.33&\phantom{-}  \textcolor{blue}{ 0.43}& 0.33& 0.39& 0.30& 0.37&\bfseries 0.39\tabularnewline
US-INF-2Q &0.22& 0.10& 0.45& 0.28&\phantom{---}  \textcolor{blue}{\bfseries  0.45}**& 0.12&0.27***\tabularnewline
\midrule
UK-EMP-1M&-0.05& 0.03&-0.51&-0.07&-0.02&-0.04&\phantom{-}  \textcolor{blue}{\bfseries 0.04}\tabularnewline
UK-EMP-3M&0.06& 0.15&-0.49& 0.23& 0.18& 0.17&\phantom{---.--}  \textcolor{blue}{\bfseries 0.27}***\tabularnewline
UK-HRS-1M&0.05& 0.00&-0.14& 0.03&-0.05& 0.00&\phantom{--.}  \textcolor{blue}{\bfseries 0.08}*\tabularnewline
UK-HRS-3M&0.01& 0.16& 0.01& 0.22&\phantom{-}  \textcolor{blue}{\bfseries  0.27}& 0.24&0.22\tabularnewline
UK-INF-1M&\phantom{-}  \textcolor{blue}{ 0.73}& 0.13& 0.32& 0.08& 0.25***&\bfseries  0.26&0.26\tabularnewline
UK-INF-3M&\phantom{-}  \textcolor{blue}{ 0.74}&-0.27&-0.03& 0.05& 0.14& 0.03&\bfseries 0.24***\tabularnewline

\toprule
\end{tabular}
\begin{tablenotes}[para,flushleft]
  Notes: This table reports out-of-sample $R^2$'s for 24 data sets and different models, either standard or introduced in the text. For macroeconomic targets (the last 14 data sets), the set of benchmark models additionally includes an autoregressive model (AR) of order 2 for quarterly data and of order 6 for monthly data. Numbers in \textbf{bold} identify the best predictive performance among tree ensembles. Numbers in \textcolor{blue}{blue} are the best performances over all models. For tree ensembles, t-test (and  \citet{dieboldmariano} tests for time series data) are performed to evaluate the statistical significance of the accuracy differential between the "standard" versions (either RF or BT) and the alternative regularization versions (SGT or Booging) . '*', '**' and '***' respectively refer to p-values below 10\%, 5\% and 1\%.
  \end{tablenotes}
  \end{threeparttable}
  \vspace*{-0.65cm}  
  \end{center}
\end{table}  

The last 6 data sets are monthly data from the UK using the newly built database from \citet{GCMS}. The 4 variables are employment growth (EMP), hours worked change (HRS), and CPI inflation (INF). Those are predicted at horizons of 1 month and 3 months. For the latter, the target is the average growth rate (or change) over the next 3 months. The underlying data set is similar in nature to \citet{mccrackenng} and transformations similar to what described above are performed. Each data set has about 260 observations (ending in 2019) and a number of predictors ranging from 250 to about 400. 

Most data sets can be considered "small data".  Those applications empirically motivates the development of SGT -- beyond that it is a surrogate tool to understand RF. The reason is simple: the choice of regularizer matters more for small $N$. SGT and RF precisely differ in that aspect. 

The models of interest are the four members of the tree ensemble quaternity, i.e., BT, RF, Booging and SGT$(\eta=0.1,\bar{H}=0.25)$. BT, CART, Lasso, and RF are tuned (which also includes $\nu$ for BT) via k-fold cross-validation.\footnote{Note that using tree-based methods on time series is appropriate with \textit{stationary} data, which is true after applying standard econometric transformations. A growing literature shows that tree ensembles are very hard beat when it comes to macroeconomic forecasting \citep{chen2019off,medeiros2019,GCLSS2018}.}  "Benchmarks" include Lasso and CART.\footnote{CART is pruned according to CV. Lasso's $\lambda$ is a also tuned by CV.} For the macro data sets, they additionally include an autoregressive model of order 2 (AR), which is often very hard to beat \citep{KLS2019}. I keep 70\% of observations for training (and optimizing hyperparameters if needed) and use the remaining 30\% to evaluate performance. For cross-sectional data sets, those observations are chosen randomly. For time series applications, the first $70\% \times N$ observations are the training set. 



Table \ref{results1} report results. First, for standard data sets, RF and BT clearly dominates, which lend empirical support for the usual regularization/model pairs. A natural first question is whether SGT can closely match the RF stellar record. The answer is yes, SGT -- with a single greedily optimized tree -- trails behind by 0 to 7 points of $R^2$ percentages. Sometimes that difference is statistically significant (like for Red Wine, a 5\% gap significant at the 1\% level), sometimes it is not (like for Boston, Fish Toxicity and Ohio Housing). Note that SGT is miles ahead of CART for every of those 10 data sets. Hence, unlike "optimal" trees \citep{bertsimas2017optimal,blanquero2020sparse}, SGT results are in the ballpark of RF rather than merely improving upon CART.

Another interesting question is whether successes are mostly driven by the choice of regularizer or rather by that of the inherent model. The answer largely depends on the case under study. Results of NBA Salary, Red Wine, and Fish Toxicity are indicative of the one deep tree being more appropriate than shallow additive trees for those data sets (RF$\succ$Booging and SGT$\succ$BT). In contrast, results for California Housing suggest that opting for diversification over slow learning is the game changer (RF$\succ$SGT and Booging$\succ$BT).




The difficulties evoked earlier for macro data sets are clearly visible, with most $R^2$'s being rather low. A first observation is that RF is \textit{never} the best among all models, nor tree ensembles. Nevertheless, it still offers highly competitive results. When it comes to US-GDP at 1 quarter ahead, all tree ensembles are alike with a small edge for those based on diversification. When it comes to unemployment rate (UR) at horizon 1, Booging supplants all models. At horizon 2, it still dominates, but is now ex-aequo with BT. For UR at horizon 1, it is interesting to notice how diversification-based tree ensembles supplant all slow learning-based variants. For inflation (INF) at horizon 1, all tree ensembles give similar performance, with SGT lagging behind in a non-statistically significant fashion. However, those performances are narrowly eclipsed by Lasso. At horizon 2, Booging improves over BT by 15\%, which is significant at the 1\% level. SGT strongly improves over RF for the 2-quarter ahead forecast (17\%, which is significant at the 5\% level) and is the best by far among tree ensembles. In terms of MSE, it is matched by CART, but SGT has a mean absolute error that is lower by 7\%. Thus, SGT generates the best 2-quarters ahead inflation forecasts overall.

\begin{wrapfigure}{r}{0.5\textwidth}  
\begin{center} 
\vspace*{-0.95cm}  
\hspace*{-0.95cm}\includegraphics[scale=.4]{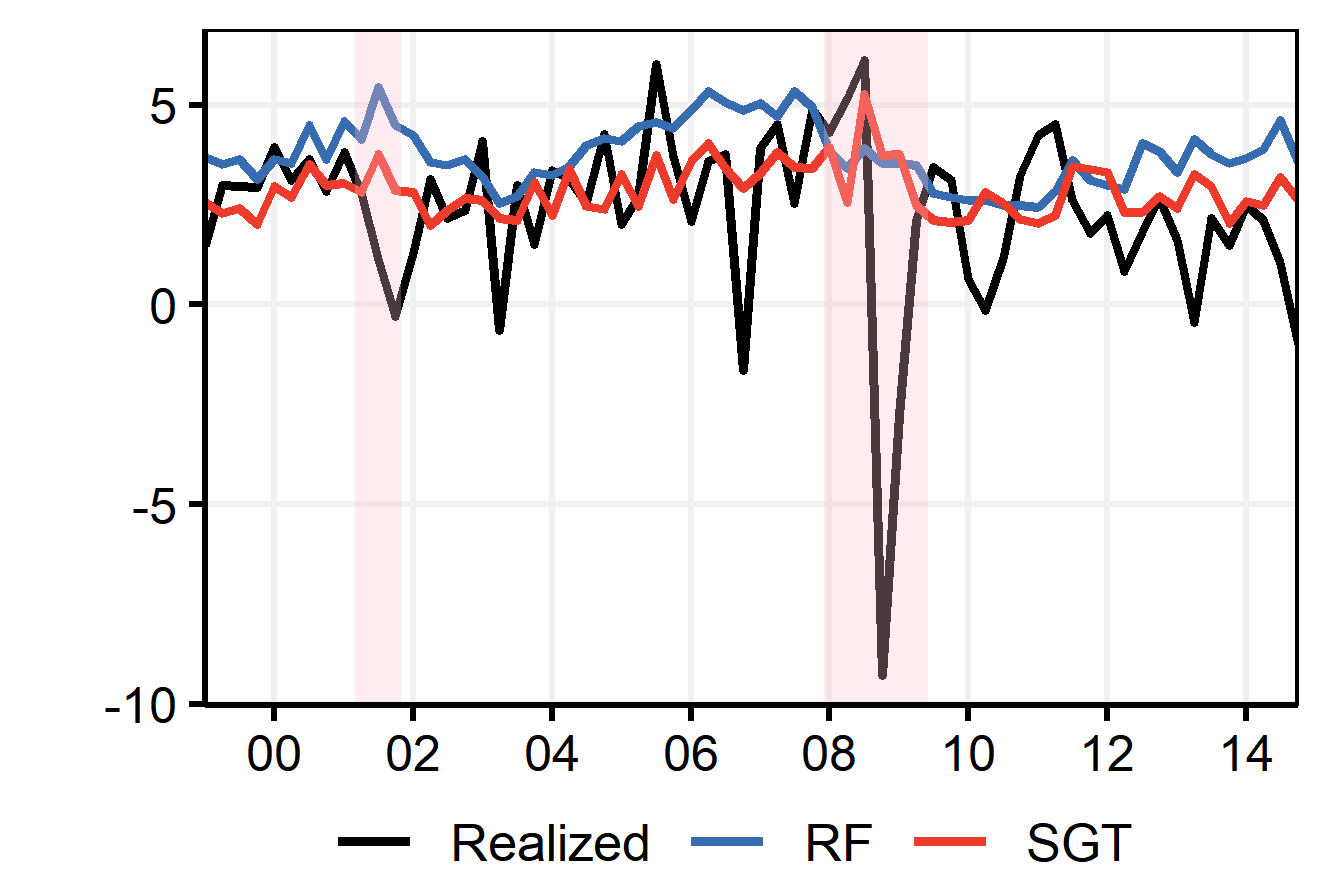}
\vspace*{-0.25cm}  
\caption{\footnotesize Comparing the forecasts from RF and SGT for US inflation (from one quarter to the next, annualized) at a horizon of 2 quarter. Pink-shaded regions are recessions.}
\label{gdph1}
\vspace*{-0.65cm}  
\end{center}
\end{wrapfigure}

Figure \ref{gdph1} investigate where SGT inflation forecasting gains originate from. At a horizon of half a year ahead, it is understood that many economic shocks cannot be captured by learning algorithms --- one example is the 2008 downward spike attributed to an unprecedented collapse of oil prices. Rather, a good inflation forecast at such a range is a reliable estimate of the lower frequency movements of inflation, which is in itself a notoriously hard problem \citet{stock2008phillips}. SGT clearly provides that. While RF mostly overshoots, SGT matches nicely the slowly decreasing then increasing trend of inflation up to 2008. Moreover, the forecast is immune to the "missing disinflation puzzle" following the Great Recession. That is, unlike many macroeconomic and econometric models \citep{coibion2015phillips,linde2019resolving}, SGT does not wrongfully predict an overly large decrease in inflation following 2008. Finally, SGT is more resilient than RF when it comes to the "missing inflation puzzle" occurring in the last years of the sample.




In the noisy environment of monthly time series, many gains are not statistically significant. There are exceptions. SGT's ($R^2$=25\%) gain over RF ($R^2$=08\%) for inflation one month ahead is statistically significant at the 5\% level.  However, this is far from being enough to capture UK inflation dynamics: SGT, BT and Booging all trail behind the more appropriate AR by a large margin.\footnote{This motivates \citet{MRF} to develop a type of RF which can accommodate more easily (among other things) for the persistence of most macroeconomic time series.} There are less dismal cases -- like that of hours worked 3 months ahead where SGT is the best model overall. This points out that SGT may be a potent tool to extract the faint signal out notoriously noisy series depicting real economic activity. Finally, Booging provides the best forecast for employment at both horizons, and hours worked at horizon 1. For EMP at the 3 months horizon, diversification clearly does better than slow learning with RF $\succ$ SGT and Booging $\succ$ BT. The edge of Booging over BT is statistically significant at the 1\% level.



\section{Conclusion}\label{sec:con}

This paper proposes a tree ensemble taxonomy which ultimately helps understand what RF is doing. To make that claim, Slow-Growing Trees are introduced. It is shown that learning a single SGT delivers results often similar to that of RF. SGT allows linking back RF to boosting theory where the latter is argued to be similar to a very high-dimensional $l_1$-regularized regression problem.  Simulations confirm that SGT and RF coincide when the true DGP is a tree. Applications to real data sets show that both SGT and Booging (additive trees regularized via diversification rather than a low learning rate) can provide gains over both BT and RF in difficult learning environments. Thus,  for tree ensembles, regularization swapping  can be beneficial.

\nocite{efron2004least}

\clearpage

\setlength\bibsep{5pt}
		
\bibliographystyle{apalike}

\setstretch{0.75}

\bibliography{/Users/UQAM/Dropbox/ref_pgc_v181204}

\begin{thebibliography}{}

\bibitem[Arik and Pfister, 2019]{arik2019tabnet}
Arik, S.~O. and Pfister, T. (2019).
\newblock Tabnet: Attentive interpretable tabular learning (2019).
\newblock {\em arXiv preprint arXiv:1908.07442}.

\bibitem[{\'A}vila et~al., 2013]{avila2013concentration}
{\'A}vila, F., Flores, E., L{\'o}pez-Gallo, F., and M{\'a}rquez, J. (2013).
\newblock Concentration indicators: Assessing the gap between aggregate and
  detailed data.
\newblock {\em IFC Bulletin}, 36:542--559.

\bibitem[Azar et~al., 2020]{azar2020labor}
Azar, J., Marinescu, I., and Steinbaum, M. (2020).
\newblock Labor market concentration.
\newblock {\em Journal of Human Resources}, pages 1218--9914R1.

\bibitem[Bartlett et~al., 2020]{bartlett2020benign}
Bartlett, P.~L., Long, P.~M., Lugosi, G., and Tsigler, A. (2020).
\newblock Benign overfitting in linear regression.
\newblock {\em Proceedings of the National Academy of Sciences}.

\bibitem[Belkin et~al., 2019]{belkin2019reconciling}
Belkin, M., Hsu, D., Ma, S., and Mandal, S. (2019).
\newblock Reconciling modern machine-learning practice and the classical
  bias--variance trade-off.
\newblock {\em Proceedings of the National Academy of Sciences},
  116(32):15849--15854.

\bibitem[Bertsimas and Dunn, 2017]{bertsimas2017optimal}
Bertsimas, D. and Dunn, J. (2017).
\newblock Optimal classification trees.
\newblock {\em Machine Learning}, 106(7):1039--1082.

\bibitem[Blanquero et~al., 2020]{blanquero2020sparse}
Blanquero, R., Carrizosa, E., Molero-R{\i}o, C., and Morales, D.~R. (2020).
\newblock On sparse optimal regression trees.
\newblock Technical report, Technical report, IMUS, Sevilla, Spain,
  https://www. researchgate. net~….

\bibitem[Breiman, 2001]{breiman2001}
Breiman, L. (2001).
\newblock Random forests.
\newblock {\em Machine learning}, 45(1):5--32.

\bibitem[Breiman et~al., 1984]{breiman1984classification}
Breiman, L., Friedman, J., Stone, C.~J., and Olshen, R.~A. (1984).
\newblock {\em Classification and regression trees}.
\newblock CRC press.

\bibitem[B{\"u}hlmann et~al., 2007]{buhlmann2007boosting}
B{\"u}hlmann, P., Hothorn, T., et~al. (2007).
\newblock Boosting algorithms: Regularization, prediction and model fitting.
\newblock {\em Statistical Science}, 22(4):477--505.

\bibitem[B{\"u}hlmann and Yu, 2003]{buhlmann2003boosting}
B{\"u}hlmann, P. and Yu, B. (2003).
\newblock Boosting with the l 2 loss: regression and classification.
\newblock {\em Journal of the American Statistical Association},
  98(462):324--339.

\bibitem[B{\"u}hlmann et~al., 2002]{buhlmann2002analyzing}
B{\"u}hlmann, P., Yu, B., et~al. (2002).
\newblock Analyzing bagging.
\newblock {\em The Annals of Statistics}, 30(4):927--961.

\bibitem[Chen et~al., 2019]{chen2019off}
Chen, J.~C., Dunn, A., Hood, K.~K., Driessen, A., and Batch, A. (2019).
\newblock Off to the races: A comparison of machine learning and alternative
  data for predicting economic indicators.
\newblock In {\em Big Data for 21st Century Economic Statistics}. University of
  Chicago Press.

\bibitem[Coibion and Gorodnichenko, 2015]{coibion2015phillips}
Coibion, O. and Gorodnichenko, Y. (2015).
\newblock Is the phillips curve alive and well after all? inflation
  expectations and the missing disinflation.
\newblock {\em American Economic Journal: Macroeconomics}, 7(1):197--232.

\bibitem[Diebold and Mariano, 2002]{dieboldmariano}
Diebold, F.~X. and Mariano, R.~S. (2002).
\newblock Comparing predictive accuracy.
\newblock {\em Journal of Business \& economic statistics}, 20(1):134--144.

\bibitem[Efron et~al., 2004]{efron2004least}
Efron, B., Hastie, T., Johnstone, I., Tibshirani, R., et~al. (2004).
\newblock Least angle regression.
\newblock {\em The Annals of statistics}, 32(2):407--499.

\bibitem[Freund et~al., 1996]{freund1996experiments}
Freund, Y., Schapire, R.~E., et~al. (1996).
\newblock Experiments with a new boosting algorithm.
\newblock In {\em icml}, volume~96, pages 148--156. Citeseer.

\bibitem[Friedman et~al., 2001]{ESL}
Friedman, J., Hastie, T., and Tibshirani, R. (2001).
\newblock {\em The elements of statistical learning}, volume~1.
\newblock Springer series in statistics New York, NY, USA:.

\bibitem[Friedman, 1991]{MARS}
Friedman, J.~H. (1991).
\newblock Multivariate adaptive regression splines.
\newblock {\em The annals of statistics}, pages 1--67.

\bibitem[Friedman, 2001]{friedman2001}
Friedman, J.~H. (2001).
\newblock Greedy function approximation: a gradient boosting machine.
\newblock {\em Annals of statistics}, pages 1189--1232.

\bibitem[Friedman, 2002]{friedman2002}
Friedman, J.~H. (2002).
\newblock Stochastic gradient boosting.
\newblock {\em Computational statistics \& data analysis}, 38(4):367--378.

\bibitem[FTC/DOJ, 2010]{DOJ}
FTC/DOJ (2010).
\newblock Horizontal merger guidelines.

\bibitem[Goodfellow et~al., 2016]{deeplearning}
Goodfellow, I., Bengio, Y., Courville, A., and Bengio, Y. (2016).
\newblock {\em Deep learning}, volume~1.
\newblock MIT press Cambridge.

\bibitem[Goulet~Coulombe, 2020a]{MRF}
Goulet~Coulombe, P. (2020a).
\newblock The macroeconomy as a random forest.
\newblock {\em arXiv preprint arXiv:2006.12724}.

\bibitem[Goulet~Coulombe, 2020b]{MSoRF}
Goulet~Coulombe, P. (2020b).
\newblock To bag is to prune.
\newblock {\em arXiv preprint arXiv:2008.07063}.

\bibitem[Goulet~Coulombe et~al., 2019]{GCLSS2018}
Goulet~Coulombe, P., Leroux, M., Stevanovic, D., Surprenant, S., et~al. (2019).
\newblock How is machine learning useful for macroeconomic forecasting?
\newblock Technical report, CIRANO.

\bibitem[Goulet~Coulombe et~al., 2020]{MDTM}
Goulet~Coulombe, P., Leroux, M., Stevanovic, D., Surprenant, S., et~al. (2020).
\newblock Macroeconomic data transformations matter.
\newblock Technical report, CIRANO.

\bibitem[Goulet~Coulombe et~al., 2021]{GCMS}
Goulet~Coulombe, P., Marcellino, M., and Stevanovic, D. (2021).
\newblock Can machine learning catch the covid-19 recession?
\newblock {\em CEPR Discussion Paper No. DP15867}.

\bibitem[Gu et~al., 2020]{kellyml}
Gu, S., Kelly, B., and Xiu, D. (2020).
\newblock Empirical asset pricing via machine learning.
\newblock {\em The Review of Financial Studies}, 33(5):2223--2273.

\bibitem[Hastie et~al., 2007]{hastie2007forward}
Hastie, T., Taylor, J., Tibshirani, R., Walther, G., et~al. (2007).
\newblock Forward stagewise regression and the monotone lasso.
\newblock {\em Electronic Journal of Statistics}, 1:1--29.

\bibitem[Hu et~al., 2019]{hu2019optimal}
Hu, X., Rudin, C., and Seltzer, M. (2019).
\newblock Optimal sparse decision trees.
\newblock In {\em Advances in Neural Information Processing Systems}, pages
  7267--7275.

\bibitem[Irsoy et~al., 2012]{irsoy2012soft}
Irsoy, O., Y{\i}ld{\i}z, O.~T., and Alpayd{\i}n, E. (2012).
\newblock Soft decision trees.
\newblock In {\em Proceedings of the 21st International Conference on Pattern
  Recognition (ICPR2012)}, pages 1819--1822. IEEE.

\bibitem[Jordan and Jacobs, 1994]{jordan1994hierarchical}
Jordan, M.~I. and Jacobs, R.~A. (1994).
\newblock Hierarchical mixtures of experts and the em algorithm.
\newblock {\em Neural computation}, 6(2):181--214.

\bibitem[Kobak et~al., 2020]{kobak2020optimal}
Kobak, D., Lomond, J., and Sanchez, B. (2020).
\newblock The optimal ridge penalty for real-world high-dimensional data can be
  zero or negative due to the implicit ridge regularization.
\newblock {\em Journal of Machine Learning Research}, 21(169):1--16.

\bibitem[Kotchoni et~al., 2019]{KLS2019}
Kotchoni, R., Leroux, M., and Stevanovic, D. (2019).
\newblock Macroeconomic forecast accuracy in a data-rich environment.
\newblock {\em Journal of Applied Econometrics}, 34(7):1050--1072.

\bibitem[Lind{\'e} and Trabandt, 2019]{linde2019resolving}
Lind{\'e}, J. and Trabandt, M. (2019).
\newblock Resolving the missing deflation puzzle.

\bibitem[Mason et~al., 2000]{mason2000boosting}
Mason, L., Baxter, J., Bartlett, P.~L., and Frean, M.~R. (2000).
\newblock Boosting algorithms as gradient descent.
\newblock In {\em Advances in neural information processing systems}, pages
  512--518.

\bibitem[McCracken and Ng, 2020]{mccracken2020fred}
McCracken, M. and Ng, S. (2020).
\newblock Fred-qd: A quarterly database for macroeconomic research.
\newblock Technical report, National Bureau of Economic Research.

\bibitem[McCracken and Ng, 2016]{mccrackenng}
McCracken, M.~W. and Ng, S. (2016).
\newblock Fred-md: A monthly database for macroeconomic research.
\newblock {\em Journal of Business \& Economic Statistics}, 34(4):574--589.

\bibitem[Medeiros et~al., 2019]{medeiros2019}
Medeiros, M.~C., Vasconcelos, G.~F., Veiga, {\'A}., and Zilberman, E. (2019).
\newblock Forecasting inflation in a data-rich environment: the benefits of
  machine learning methods.
\newblock {\em Journal of Business \& Economic Statistics},
  (just-accepted):1--45.

\bibitem[Molnar, 2019]{molnar2019interpretable}
Molnar, C. (2019).
\newblock {\em Interpretable machine learning}.
\newblock Lulu.com.

\bibitem[Norouzi et~al., 2015]{norouzi2015efficient}
Norouzi, M., Collins, M., Johnson, M.~A., Fleet, D.~J., and Kohli, P. (2015).
\newblock Efficient non-greedy optimization of decision trees.
\newblock In {\em Advances in neural information processing systems}, pages
  1729--1737.

\bibitem[Rosset et~al., 2004]{rosset2004boosting}
Rosset, S., Zhu, J., and Hastie, T. (2004).
\newblock Boosting as a regularized path to a maximum margin classifier.
\newblock {\em Journal of Machine Learning Research}, 5(Aug):941--973.

\bibitem[Schapire, 2013]{schapire2013explaining}
Schapire, R.~E. (2013).
\newblock Explaining adaboost.
\newblock In {\em Empirical inference}, pages 37--52. Springer.

\bibitem[Stock and Watson, 2008]{stock2008phillips}
Stock, J.~H. and Watson, M.~W. (2008).
\newblock Phillips curve inflation forecasts.
\newblock Technical report, National Bureau of Economic Research.

\bibitem[Tibshirani, 1996]{tibshirani1996}
Tibshirani, R. (1996).
\newblock Regression shrinkage and selection via the lasso.
\newblock {\em Journal of the Royal Statistical Society. Series B
  (Methodological)}, pages 267--288.

\bibitem[Wyner et~al., 2017]{wyner2017explaining}
Wyner, A.~J., Olson, M., Bleich, J., and Mease, D. (2017).
\newblock Explaining the success of adaboost and random forests as
  interpolating classifiers.
\newblock {\em The Journal of Machine Learning Research}, 18(1):1558--1590.

\bibitem[Zhang and Yu, 2005]{zhang2005boosting}
Zhang, T. and Yu, B. (2005).
\newblock Boosting with early stopping: Convergence and consistency.
\newblock {\em The Annals of Statistics}, 33(4):1538--1579.

\end{thebibliography}

\clearpage

\end{document}